\pdfoutput=1

\documentclass[11pt]{article}

\usepackage[preprint]{acl}

\usepackage{times}
\usepackage{latexsym}

\usepackage{graphicx} 

\usepackage[T1]{fontenc}

\usepackage[utf8]{inputenc}

\usepackage{microtype}
\usepackage{booktabs}
\usepackage{inconsolata}
\usepackage{algorithm}
\usepackage{algorithmic}
\usepackage{paralist}

\usepackage{tcolorbox}
\usepackage{adjustbox}

\usepackage{color}

\definecolor{nmgray}{RGB}{229,229,229}
\title{NUS-Emo at SemEval-2024 Task 3: 
Instruction-Tuning LLM for Multimodal Emotion-Cause Analysis in Conversations
}

\author{
  Meng Luo$^1$\thanks{Equal contributions.} \quad
  Han Zhang$^{2*}$ \quad  Shengqiong Wu$^1$ \quad Bobo Li$^3$ \quad
  Hong Han$^2$ \quad Hao Fei$^1$\thanks{Corresponding author.}
  \\
  $^1$National University of Singapore  \quad 
  $^2$Xidian University  \quad 
  $^3$Wuhan University
  \\
  \texttt{mluo@u.nus.edu} \quad \texttt{zhanghanxd@stu.xidian.edu.cn} \quad \texttt{swu@u.nus.edu} \quad \\ 
  \texttt{boboli@whu.edu.cn} \quad \texttt{hanh@mail.xidian.edu.cn} \quad
  \texttt{haofei37@nus.edu.sg}
}

\begin{document}
\maketitle
\begin{abstract}
This paper describes the architecture of our system developed for Task 3 of SemEval-2024: Multimodal Emotion-Cause Analysis in Conversations. 
Our project targets the challenges of subtask 2, dedicated to Multimodal Emotion-Cause Pair Extraction with Emotion Category (MECPE-Cat), and constructs a dual-component system tailored to the unique challenges of this task. 
We divide the task into two subtasks: emotion recognition in conversation (ERC) and emotion-cause pair extraction (ECPE). 
To address these subtasks, we capitalize on the abilities of Large Language Models (LLMs), which have consistently demonstrated state-of-the-art performance across various natural language processing tasks and domains. 
Most importantly, we design an approach of emotion-cause-aware instruction-tuning for LLMs, to enhance the perception of the emotions with their corresponding causal rationales.
Our method enables us to adeptly navigate the complexities of MECPE-Cat, achieving a weighted average 34.71\% F1 score of the task,
and securing the \textbf{2$^{nd}$} rank on the leaderboard.\footnote{\url{https://nustm.github.io/SemEval-2024_ECAC/}}
The code and metadata to reproduce our experiments are all made publicly available.\footnote{\url{https://github.com/zhanghanXD/NUS-Emo-at-SemEval-2024-Task3}}
\end{abstract}

\section{Introduction}

Emotion cause analysis is a critical component of human communication and decision-making, offering substantial applications across diverse fields. 
It enables a deeper and more detailed understanding of sentiments. 
The introduction of emotion-cause analysis in textual conversations by \citet{poria2021recognizing} has paved the way for advancements in understanding emotional dynamics within dialogues. 
However, textual analysis alone does not fully capture the complexity of human emotional expression, as emotions and their causes are often conveyed through a blend of modalities \cite{hazarika2018icon,Wu00C23,0001LZZC23}. 
Subtask 2 of SemEval-2024 Task 3, referred to as MECPE-Cat, seeks to expand this analysis into the multimodal domain, focusing on English-language conversations. 
The task draws inspiration from the seminal work of \citet{wang2021multimodal}, which sets out to jointly extract emotions and their corresponding causes from conversations across multiple modalities, including text, audio, and video, and it also encompasses the identification of the corresponding emotion category for each emotion-cause pair.

\begin{figure*}[htbp]
\centering
\includegraphics[width=\textwidth]{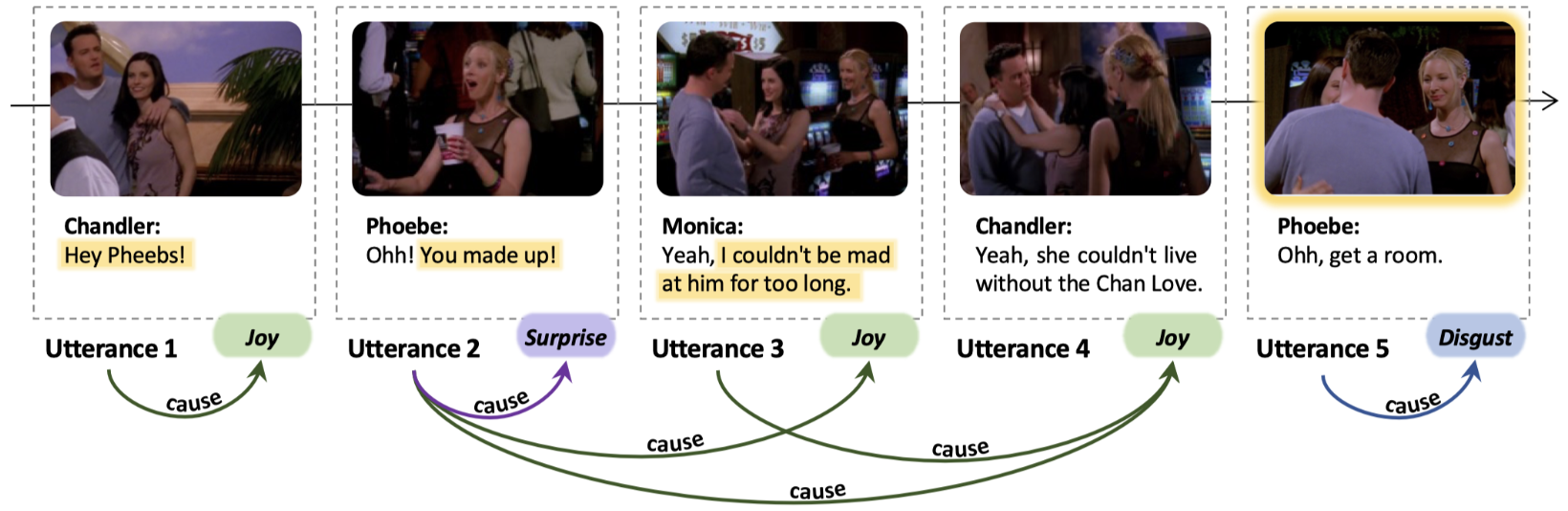}
\caption{An example of an official task and annotated dataset. Each arc points from the cause utterance to the emotional triggers. The cause spans have been highlighted in yellow. Background: Chandler and his girlfriend Monica walked into the casino (they had a quarrel earlier but made up soon), and then started a conversation with Phoebe.}
\label{fig:figure1}
\end{figure*}

In our system, we leverage LLMs such as GPT-3 \cite{brown2020language}, Flan-T5 \cite{chung2022scaling}, and GLM \cite{du2021glm} known for their exceptional performance in various natural language processing tasks. 
We employ parameter-efficient fine-tuning, specifically LoRA \cite{hu2021lora}, to efficiently fine-tune LLMs, enhancing their performance with minimal computational overhead. 
Additionally, we harness emotion-cause-aware prompt-based learning and instruction-tuning to enhance model performance such that the LLMs can more accurately perceive the emotions with their corresponding causal rationales. 
Prompt-based learning guides LLMs to generate contextually relevant outputs, while instruction-fine-tuning models for our specific tasks by improving their response to explicit instructions.

In this paper, we investigate the optimal LLM for the MECPE-Cat task, selecting ChatGLM based on its superior zero-shot performance. 
We further refine ChatGLM through instruction-tuning, using carefully crafted prompts to enhance its task-specific accuracy. 
Our fine-tuned model achieves the second-highest score on the official test set for subtask 2, with a weighted average of 34.71\% F1, underscoring the effectiveness of our approach. 
We also discuss the current limitations of our model and methodology, alongside directions for future research and improvement. 
We will release our codes and resources mentioned in this paper to facilitate relevant research.

\section{Background}
\subsection{Task and Dataset Description}

The SemEval-2024 Task 3 is based on the multimodal conversational emotion-cause dataset, Emotion-Cause-in-Friends \citep[ECF; ][]{wang2021multimodal}, by choosing a multimodal dataset MELD \cite{poria2018meld} as the data source and further annotating the corresponding causes for the given emotion annotations. 
The ECF dataset contains 9,794 emotion-cause pairs, covering three modalities. 
The subtask 2 is to extract all emotion-cause pairs in a given conversation under three modalities, where each pair contains an emotion utterance along with its emotion category and a cause utterance, e.g., (U3\_Joy, U2), which means that the speaker’s joy emotion in utterance 3 is triggered by the cause from utterance 2. 
Figure \ref{fig:figure1} displays a real example of this task and annotated dataset. 
In this conversation, it is expected to extract a set of six utterance-level emotion-cause pairs in total, e.g., Chandler’s Joy emotion in Utterance 4 (U4 for short) is triggered by the objective cause that he and Monica had made up and Monica’s subjective opinion in U3, forming the pairs (U4\_joy, U2) and (U4\_joy, U3); The cause for Phoebe’s Disgust in U5 is the objective event that Monica and Chandler were kissing in front of her (mainly reflected in the visual modality of U5), forming the pair (U5\_disgust, U5).

\subsection{Related Work}
The exploration of ECPE within textual and conversational contexts has been approached through various methodologies, each tailored to specific task settings \cite{ChenSLW0LJ22}. \citet{cheng2023consistent} reframe the ECPE task as a process akin to engaging in a two-stage machine reading comprehension (MRC) challenge. \citet{zheng2023ecqed} expand the ECPE task to Emotion-Cause Quadruple Extraction in Dialogs (ECQED), focusing on detecting pairs of emotion-cause utterances and their types. 
They present a model utilizing a heterogeneous graph and a parallel grid tagging scheme for this purpose. In addressing the specific challenge of the MECPE-Cat task, \citet{wang2021multimodal} set a benchmark for this task by introducing two preliminary baseline systems. They utilize a heuristic approach to leverage inherent patterns in the localization of causes and emotions, alongside a deep learning strategy, MECPE-2steps, which adapts a prominent ECPE methodology for news articles to include multimodal data.

Drawing from the varied methodologies of previous work, it becomes clear that effectively solving the MECPE-Cat task demands a deep understanding of dialogue content, precise identification of conversational emotions, extraction of emotion-cause pairs, and the integration of multimodal information. 
Motivated by the strong performance of LLMs on various metrics, we opt to utilize these models to address this intricate challenge. 
Through exhaustive model evaluations and extensive prompt testing, we have showcased the practicality, superiority, and adaptability of our chosen approach.

\begin{figure}[!ht] 
\centering
\includegraphics[width=0.98\columnwidth]{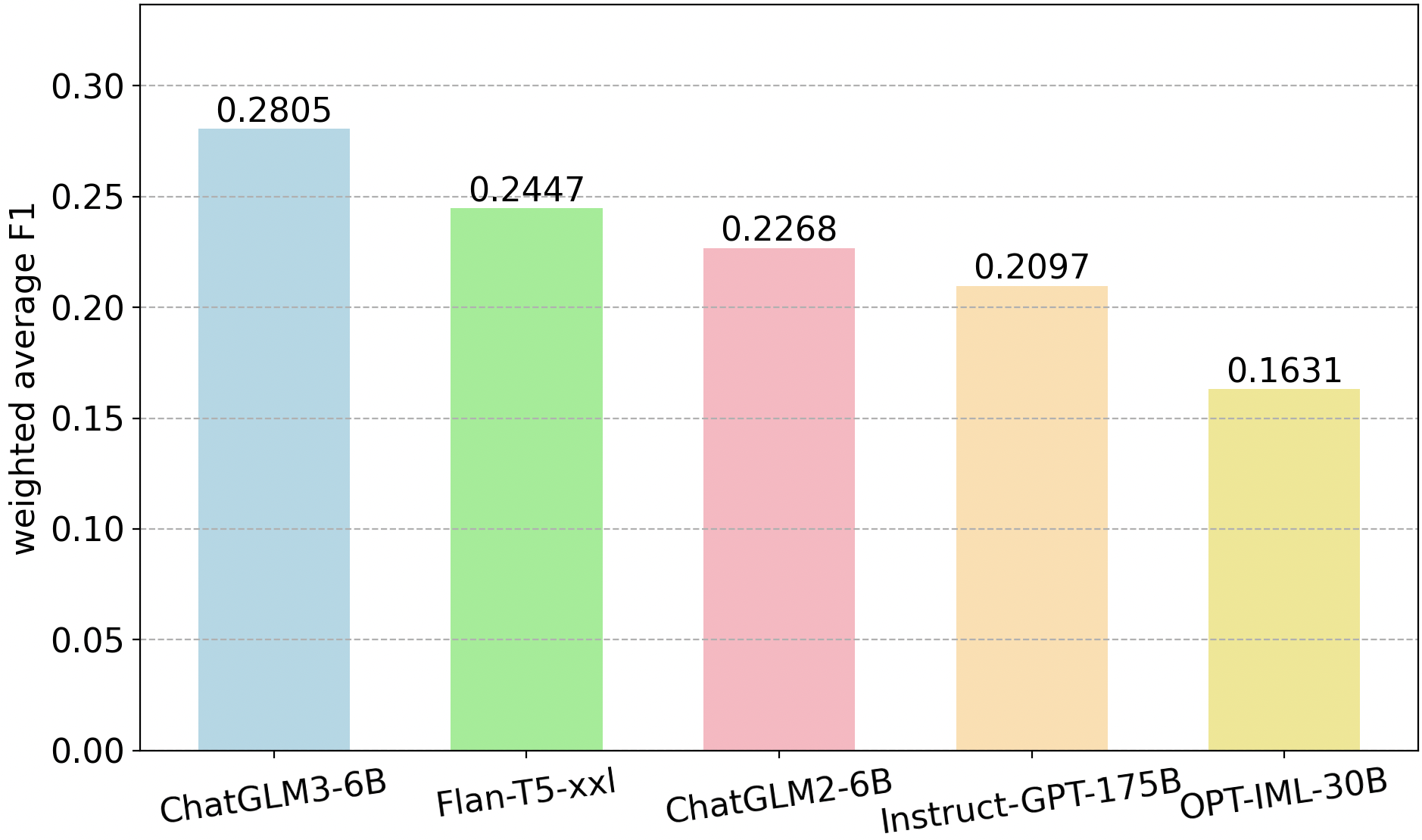} 
\caption{Zero-shot test set performance of various instruction-tuned LLMs.}
\label{fig:figure2}
\end{figure}

\section{Methodology}
In this section, we first conduct preliminary experiments to determine which LLM to select as a backbone reasoner.
We then elaborate on how we design the system and emotion-cause-aware instructions for tuning our chosen LLM.

\subsection{Pilot Study for LLM Selection}
Currently, there exists a variety of LLMs, such as OPT-IML \cite{iyer2022opt}, GPT-3, Flan-T5, and GLM. 
However, it is essential to select a model that not only performs optimally but is also the most suitable for our specific task. 
To this end, we carry out a pilot study to determine the most appropriate model selection.
For our zero-shot testing experiment, we rigorously evaluate several models, including OPT-IML\footnote{OPT-IML-30B, max version with 30B, \url{https://huggingface.co/facebook/opt-iml-30b}}, Instruct-GPT\footnote{Instruct-GPT-175B, an advanced version of the GPT-3.5.} \cite{ouyang2022training}, Flan-T5\footnote{Flan-T5-xxl, with 11B, \url{https://huggingface.co/google/flan-t5-xxl}}, alongside the ChatGLM models, to identify the most effective tool for this specific task. 
We customize instructions for each model's specific tuning style, recognizing that a single set of instructions does not suit all models effectively. We also embed expected output labels within these instructions to secure precise responses from each model. 
Figure \ref{fig:figure2} depicts the zero-shot performance of these models. 
The ChatGLM\footnote{ChatGLM, 3rd version with 6B, \url{https://github.com/THUDM/ChatGLM3}.} LLM is ultimately selected based on its superior performance in these tests. 
This selection is informed not merely by the innovative features or the advanced training methodologies of ChatGLM but by empirical evidence of its exceptional zero-shot performance among the models considered.

\begin{figure}[!t] 
\centering
\includegraphics[width=0.98\columnwidth]{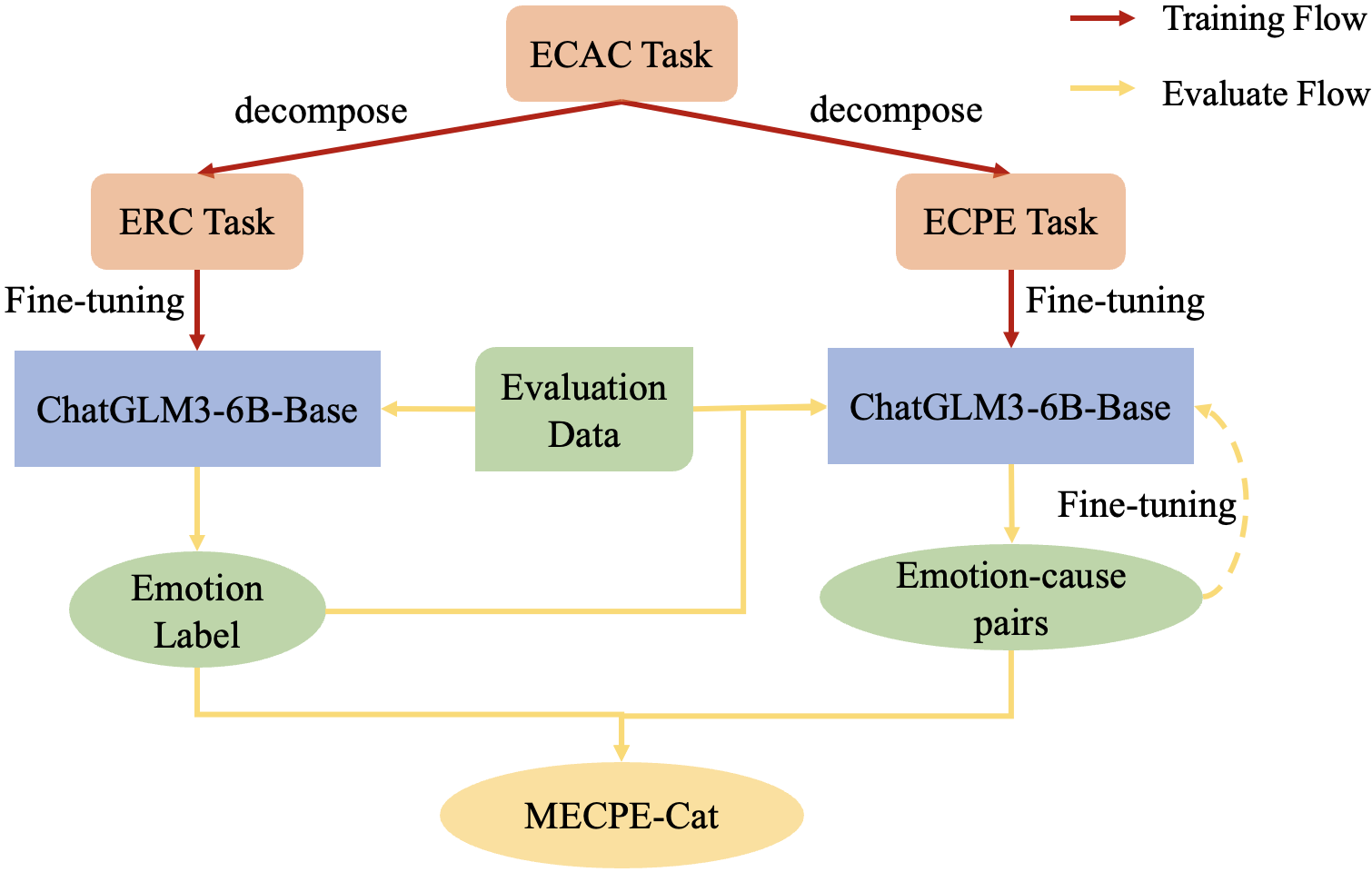}
\caption{Proposed method workflow for the MECPE-Cat task.}
\label{fig:figure3}
\end{figure}

\begin{figure}[!t] 
\centering
\includegraphics[width=0.98\columnwidth]{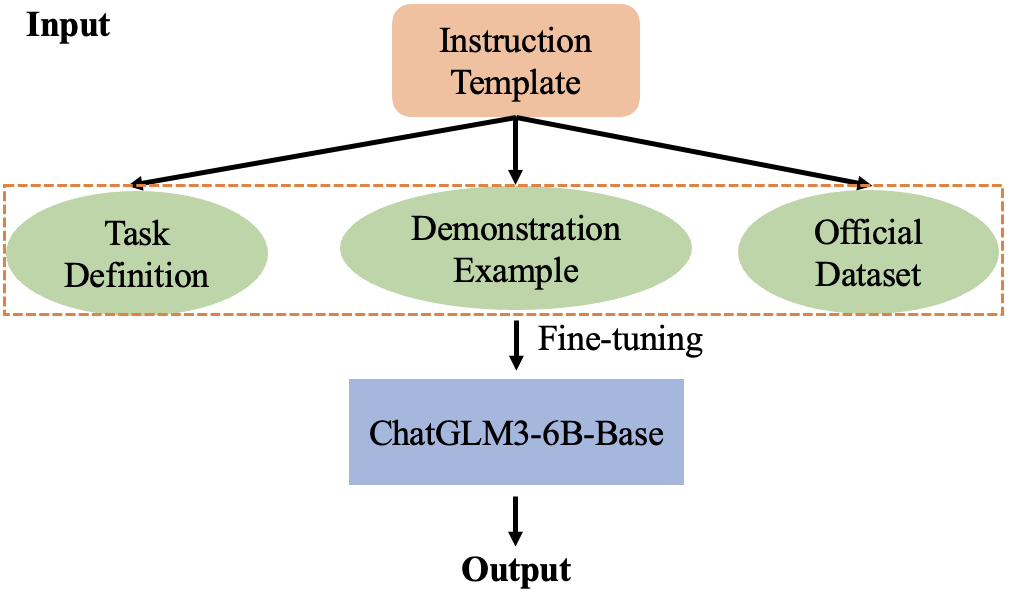} 
\caption{The construction of the instruction template and the flow of model input and output.}
\label{fig:figure4}
\end{figure}

\subsection{Multimodal Feature Encoding}

Given that the inputs for our task incorporate multimodal signals, including visual information to assist in more accurate emotion recognition, it is imperative to fully leverage the non-textual modal information. 
However, our LLM backbone does not natively support the direct inclusion of non-textual modal signals. 
To address this, we consider employing ImageBind \cite{girdhar2023imagebind} for encoding the multimodal portion of input information, owing to its robust multimodal alignment capabilities and visual perception proficiency. 
Subsequently, we concatenate the multimodal representations with other textual embeddings before feeding them into the LLM.

\subsection{Constructing Emotion-Cause-aware Instructions for LLM Tuning}

Figure \ref{fig:figure3} first illustrates the workflow of our proposed framework. 
Initially, we fine-tune the model on the ERC task. Following this, we incorporate the predicted emotion labels into each utterance, setting the stage for the ECPE task execution. 
Subsequently, we employ the model, now fine-tuned with data labeled with emotion tags, to perform inference on the MECPE-Cat task, yielding an initial set of emotion-cause pairs. 
These preliminary results are then reintegrated into the original training dataset for a second round of fine-tuning, culminating in the refinement of our model to produce the final set of emotion-cause pairs.

\begin{tcolorbox} 
{
\textit{\bf Task Definition:}\\
\textit{``You're an expert in sentiment analysis and emotion cause identification. Below is a conversation containing multiple utterances from different speakers, along with the corresponding emotion label for each utterance. Your task is to identify the indices of the candidate utterances that elicited the emotion in the target utterance.''}\\

\textit{\bf Input conversation:}\\
\textit{1\_joy. Chandler: Hey Pheebs!}\\
\textit{2\_surprise. Phoebe: Ohh! You made up!}\\
\textit{3\_joy. Monica: Yeah, I couldn't be mad at him for too long.}\\
\textit{4\_joy. Chandler: Yeah, she couldn't live without the Chan Love.}\\
\textit{5\_disgust. Phoebe: Ohh, get a room.}\\

\textit{\bf Candidate utterances:}\\
\textit{1\_joy. Chandler: Hey Pheebs!}\\
\textit{2\_surprise. Phoebe: Ohh! You made up!}\\
\textit{3\_joy. Monica: Yeah, I couldn't be mad at him for too long.}\\

\textit{\bf Target utterance:}\\
\textit{4\_joy. Chandler: Yeah, she couldn't live without the Chan Love.}\\

\textit{\bf Question:}\\
\textit{The emotion-cause indices of the target utterance are:}\\
\textcolor{red}{[LLM output]}
}
\end{tcolorbox}

To enhance the perception of identifying emotion-cause pairs and mitigate the task's inherent complexity and potential confusion, we design the template for producing emotion-cause-aware instructions to guide the model.
Figure \ref{fig:figure4} illustrates the construction of the instruction template, which encompasses the task definition, a demonstration example, and the dataset for which the model is expected to predict outcomes. 
This structured approach not only simplifies the task's complexity for the model but also aligns the model's processing capabilities with the requirements of accurately identifying emotion-cause pairs in conversations.
In the above box we showcase a real example.

\section{Experiments}

This section will quantify the effectiveness of our systems via experiments and also show more analyses to gain more observations.

\subsection{Implementation}
\label{sec:appendixA}

The hyperparameter of our system used to achieve the highest
weighted average F1 score on the subtask 2 is listed in \ref{tab:hyperparameters}. 
The ChatGLM model was fine-tuned using a learning rate of 1e-4 with LoRA-specific configurations including a rank of 8, alpha value of 32, and a dropout rate of 0.1. 
The training was conducted with a maximum instruction length of 2048 tokens and an output length limited to 128 tokens, using a batch size of 1. 
We used a single gradient accumulation step across 2 training epochs. 
These parameters were meticulously selected to optimize our model's performance.

\begin{table}[h]
\centering
\begin{tabular}{@{}lc@{}}
\toprule
Hyperparameter          & Value \\ \midrule
Learning rate           & 1e-4  \\
LoRA rank               & 8  \\
LoRA alpha              & 32  \\
LoRA dropout            & 0.1  \\
Max instruction length  & 2048  \\
Max output length       & 128     \\
Batch size              & 1     \\
Gradient accumulation steps & 1     \\
Epochs                  & 2      \\ \bottomrule
\end{tabular}
\caption{Hyperparameter used for the best performing model.}
\label{tab:hyperparameters}
\end{table}

\subsection{Evaluating Template Designing}
In constructing the instruction dataset for tuning LLMs, we systematically transform each dialogue in the dataset into training samples by embedding them into a fixed template as described above. 
The data source for this transformation is the officially provided ECF dataset, which comprises 13,619 utterances. 
Consequently, we constructed a total of 13,619 templates based on this dataset, each tailored to facilitate the model's learning and application of emotion-cause-aware instructions.

We here perform an ablation study on the contributions of each part of the instructions we designed for the task.
We derive three variants:
\begin{compactitem}
  \item \textbf{Only Task Definition:} Compared to the zero-shot paradigm, this condition offers a more detailed and precise description of the task.
  
  \item \textbf{Task + Example:} We provide a demonstrative example to clearly show the expected outcome in a real-world dialogue, offering the model a practical reference for task execution
  
  \item \textbf{Task + Example + Candidate utterances:} This design simplifies the task by introducing 'candidate utterances,' enabling the model to analyze emotion-cause pairs sentence by sentence, rather than across entire dialogues, and pinpoint the specific causes of emotions from the preceding content.
\end{compactitem}
Table \ref{tab:table1} demonstrates the comparative performance of these diverse templates.
We see that different components of the instruction templates show clear influences, such as task definition, example demonstration, and candidate utterances.
Thus, we apply all these components into our instruction templates.

\begin{table}[!t]
  \centering
  \begin{tabular}{lc}
    \toprule
    Condition & F1 Score \\
    \hline
    Only Task Definition & 0.2981 \\
    Task + Example & 0.3124 \\
    Task + Example + Candidate & \textbf{0.3207} \\
    \bottomrule
  \end{tabular}
  \caption{
  Performance using different templates for constructing instruction tuning.
  }
  \label{tab:table1}
\end{table}

\subsection{Instruction-tuning LLM}
For our experiments, we adopt a meticulous fine-tuning process for the ChatGLM. 
We set a learning rate of 1e-4, aiming for a balance between rapid convergence and maintaining the model's ability to adapt without overfitting. 
We leverage the LoRA technique with a rank of 8 and alpha of 32 to introduce task-adaptive parameters without bloating the model size, alongside a dropout rate of 0.1 to prevent overfitting.
The model processed inputs with a max sequence length of 2048 tokens, accommodating the depth of context required for our task, while the outputs are capped at 128 tokens to focus on generating concise and relevant responses. 
Both batch size and gradient accumulation steps are set to 1, tailored to our computational resources while ensuring effective backpropagation. 
This configuration, selected after careful evaluation of various setups, is instrumental in fine-tuning the ChatGLM model to achieve the best performance on our task.

Our experiments capitalize on the robust computational capabilities provided by NVIDIA A800-SXM GPUs, each boasting 80 GB of VRAM, to ensure sufficient resources are available to train large language models. 
This fine-tuning process is facilitated using a customized script derived from the Hugging Face Transformers framework, chosen for its extensive support of transformer models and seamless integration with our setup, thereby enabling us to leverage advanced hardware capabilities while utilizing a leading-edge software environment for our model's optimization.

\subsection{Task Decomposition}
We decompose the MECPE-Cat task into ERC and ECPE phases to strategically alleviate its complexity. 
This division offered a two-fold advantage: firstly, it distills the task into clearer, more focused components, facilitating a more straightforward understanding and execution of the model. 
Secondly, by leveraging emotion labels obtained from the ERC phase during the ECPE phase, we enhance the model's capability to pinpoint emotion-cause pairs with greater accuracy. Tabel \ref{tab:experiment_results} showcases incremental improvements in weighted average F1 scores across three distinct setups. 
This progression underscores the dual benefits of our approach: simplifying the task's complexity for the model and enriching the ECPE phase with contextual emotion labels, thereby optimizing the extraction of emotion-cause pairs.

\begin{table}[h!]
  \centering
  \begin{tabular}{lc}
    \toprule
    Methods & F1 Score \\
    \hline
    Single Stage & 0.3207 \\
    Two Independent Stages & 0.3288 \\
    ECPE with Emotion Labels  & \textbf{0.3396} \\
    \bottomrule
  \end{tabular}
  \caption{Comparison of weighted average F1 Scores under different methods.}
  \label{tab:experiment_results}
\end{table}

\subsection{Data Augmentation}
We find that augmenting the training dataset with trial data significantly enhanced model accuracy, achieving a high weighted average F1 score of 0.3416, as shown in Table \ref{tab:model_performance}. 
Furthermore, we employ a trick by incorporating the model's inference results on the ECPE task back into the training dataset for an additional round of fine-tuning.
This iterative fine-tuning strategy yielded a further improvement in our test data performance. 
These enhancements demonstrate the efficacy of not only expanding the training dataset but also utilizing the model's own outputs to refine its accuracy.

\begin{table}[h!]
  \centering
  \begin{tabular}{lcccc}
    \toprule
    Data  & Epoch 1 & Epoch 2 & Epoch 3 \\
    \hline
    Train & 0.3390 & 0.3396 & 0.3393 \\
    Train + Trial & 0.3404 & 0.3410 & 0.3406 \\
    Iterative Train & 0.3408 & \textbf{0.3416} & 0.3411 \\
    \bottomrule
  \end{tabular}
  \caption{Comparison of weighted average F1 Scores across different training data and epochs.}
  \label{tab:model_performance}
\end{table}

\subsection{Multimodal Integration}
To assess the impact of multimodal information on our model's performance, we adopt a methodological approach that harnessed GPT-4V \citet{achiam2023gpt} for extracting insights from modalities beyond text. 
Specifically, we enrich the instruction template with ``video description of target utterance'' derived from GPT-4V, presenting it as supplementary information to guide the model. 
This strategic integration of multimodal data leads to an improvement in the model's F1 score, as shown in Table \ref{tab:performance_comparison}, which validates the utility of multimodal information in providing richer contextual understanding.

\begin{table}[h!]
  \centering
\setlength{\tabcolsep}{5mm}
  \begin{tabular}{lc}
    \toprule
    Information & F1 Score \\
    \midrule
    Text & 0.3416 \\
    Text + Video & \textbf{0.3471} \\
    \bottomrule
  \end{tabular}
  \caption{Comparison of weighted average F1 Scores
between pure text and multimodal information.}
  \label{tab:performance_comparison}
\end{table}

\section{Conclusion}

In this work, we explore the LLMs for solving the Multimodal Emotion-Cause Pair Extraction with Emotion Category (MECPE-Cat) task.
Through a pilot study, we first select an LLM, ChatGLM, that assists in achieving optimal task performance.
The backbone ChatGLM receives textual dialogue, and also perceives the multimodal information via the ImageBind vision encoder.
Lastly, we devise an emotion-cause-aware instruction-tuning mechanism for updating LLMs, which enhances the perception of the emotions with their corresponding causal rationales. 
Our system achieves a weighted average F1 score of 34.71\%, securing second place on the MECPE-Cat leaderboard. 
\nocite{FeiRZJ23,zhang2023instruction,lester2021power,liu2023pre,wolf2020transformers,zeng2022glm,li2023revisiting,chai2022prompt,wu2023next,wu2024imagine,zhang2024context,zhao2023constructing,li2023revisiting,qu2023layoutllm,li2022diaasq,fei2023reasoning,zhang2024vpgtrans,fei2020latent}

\section*{Acknowledgements}

This work is sponsored by CCF-Baidu Open Fund.

\bibliography{custom}

\end{document}